\documentclass[10pt,twocolumn,letterpaper]{article}

\newcommand{\etal}{{\em et al\,.}}       
\newcommand{\eg}{{\em e.g.}}           
\newcommand{\ie}{{\em i.e.}}           

\usepackage{iccv}
\usepackage{times}
\usepackage{epsfig}
\usepackage{graphicx}
\usepackage{amsmath}
\usepackage{amssymb}
\usepackage{mathtools}
\usepackage[ruled,linesnumbered]{algorithm2e}
\usepackage{booktabs}
\usepackage{multirow}

\newtheorem{theorem}{Theorem}

\usepackage[pagebackref=true,breaklinks=true,letterpaper=true,colorlinks,bookmarks=false]{hyperref}

 \iccvfinalcopy 

\def\iccvPaperID{6878} 

\ificcvfinal\fi

\begin{document}

\title{Improving the Generalization of Meta-learning on Unseen Domains\\ via Adversarial Shift}

\author{Pinzhuo Tian\\
Nanjing University\\
{\tt\small tianpinzhuo@gmail.com}
\and
Yang Gao\\
Nanjing University\\
{\tt\small gaoy@nju.edu.cn}
}

\maketitle
\ificcvfinal\thispagestyle{empty}\fi

\begin{abstract}
Meta-learning provides a promising way for learning
to efficiently learn and achieves great success in many applications. However,
most meta-learning literature focuses on dealing with tasks
from a same domain, making it brittle to generalize to tasks
from the other unseen domains. In this work, we address
this problem by simulating tasks from the other unseen domains
to improve the generalization and robustness of meta-learning
method. Specifically, we propose a model-agnostic shift layer
to learn how to simulate the domain shift and generate pseudo tasks, and develop a new adversarial learning-to-learn mechanism to train it. Based on the pseudo tasks, the meta-learning model can learn cross-domain meta-knowledge, which can generalize well on unseen domains. We conduct extensive experiments under
the domain generalization setting. Experimental results demonstrate that the
proposed shift layer is applicable to various meta-learning frameworks. Moreover, our method also leads to state-of-the-art performance on different cross-domain
few-shot classification benchmarks and produces good results on cross-domain few-shot
regression.
\end{abstract}
\section{Introduction}
Learning quickly is a hallmark of human intelligence, even a child can recognize objects from a few examples. Fortunately, meta-learning provides a promising strategy for enabling efficient learning from a few supervised information, and achieves great success in many fields~\cite{DBLP:conf/icml/RakellyZFLQ19,DBLP:conf/acl/QianY19}, especially in few-shot learning~\cite{DBLP:journals/corr/abs-2004-05439,DBLP:journals/csur/WangYKN20}. However, compared with humans who can easily utilize experience from a seen environment (or domain) to help efficiently learn tasks from other unseen domains, most meta-learning models thus far have focused on the situation where
all the tasks are from a same domain. The talent of generalizing the experience to unseen domains is still a challenge for recent meta-learning methods.

Moreover, the ability of meta-learning to generalize to unseen domains is also critical in practice, due to many settings where meta-learning is applied for essentially referring to a cross-domain problem. For example, it's impossible to construct large training datasets for rare classes (\eg., some rare bird species, or some diseases), and the auxiliary set for training the meta-learning model is usually from the other domains where the annotated data is easily collected. Therefore, the meta-learning method is wished to leverage the meta-knowledge from seen domain to help efficiently study in the unseen domains. 

Although there are some works had paid attention to this problem~\cite{DBLP:conf/iclr/TsengLH020,DBLP:journals/corr/abs-2010-07734,sun2020explain}, almost all the methods tailor for classification, leading to limited applications. Typically, feature-wise transformation~\cite{DBLP:conf/iclr/TsengLH020} just can be applied to metric-based meta-learning models, and~\cite{sun2020explain} shows comparable performance to~\cite{DBLP:conf/iclr/TsengLH020}. STARTUP~\cite{DBLP:journals/corr/abs-2010-07734} must use unlabeled data from the target domain. Moreover, the performances of some methods are shown to be sensitive to the degree of the domain shift, even underperform the traditional meta-learning methods, when exists a large domain discrepancy between the training and target domains~\cite{DBLP:conf/eccv/GuoCKCSSRF20}. 

Therefore, in this work, we aim to propose a \emph{model-agnostic} and \emph{domain-free} method to improve the generalization of various meta-learning frameworks on unseen domains, in the sense that it can be applied to different learning problems, and robust whatever the domain shift is small or large. Moreover, \emph{we don't need the data from the unseen domains}.
The core idea is to generate tasks from other unseen domains, and utilize these pseudo tasks combined with true tasks sampled from the source domain to elegantly disentangle how to learn domain-invariant meta-knowledge, which can improve the generalization of meta-learning on unseen domains.  
In order to achieve this goal, we propose a shift layer to learn how to simulate the domain shift and generate tasks from unseen domains. For training it, we also develop a new adversarial learning-to-learn mechanism. In this way, the meta-learning model and the shift layer can be jointly trained end-to-end.

We evaluate the proposed method with different meta-learning models on both regression and classification problems.  Experiments demonstrate that our method is model-agnostic and robust to the degree of the domain shift.

Three primary contributions of this work are followed:
\begin{itemize}
  \item We propose a shift layer to generate pseudo tasks from unseen domains. With these pseudo tasks, the meta-learning model can easily learn cross-domain meta-knowledge.
  \vspace{-1mm}
  \item We develop an adversarial learning-to-learn mechanism to help the shift layer capture how to generate appropriate tasks which benefit for improving the generalization of the meta-learning model.
  \vspace{-1mm}
  \item The experimental results show that our method can achieve state-of-the-art performance on cross-domain few-shot classification, and also
  effectively improves the generalization of various meta-learning models on unseen domains in few-shot regression.
\end{itemize}

\section{Related Work}
\subsection{Meta-learning}
Meta-learning aims to assist the learning process in the new task by studying how learning models perform on each learning task. Recent meta-learning methods can be broadly divided into three categories, metric-based, gradient-based, and model-based methods.

\textbf{Metric-based methods.} Metric-based meta-learning framework can be considered as learning to compare, and a nonparametric similarity function is designed to evaluate the
similarity between examples. For example, Matching networks~\cite{DBLP:conf/nips/VinyalsBLKW16} firstly use attention recurrent network as a feature encoder to mapping images from different classes to a common meta-feature space, and applies cosine similarity to obtain the predicted result, Prototypical networks~\cite{DBLP:conf/nips/SnellSZ17} adopt euclidean distance, and RelationNet~\cite{DBLP:conf/cvpr/SungYZXTH18} directly utilizes a deep distance metric to measure the similarity. In general, metric-based methods are simply and effective, however thus far are restricted to classification.

\textbf{Model-based methods.} In this category, meta-learning models are usually designed as a parameterized predictor to generate parameters for the new tasks. For example, Ravi \etal~\cite{DBLP:conf/iclr/RaviL17} and Santoro \etal~\cite{santoro2016meta} both used the recurrent neural network as the predictor.

\textbf{Gradient-based methods.} Gradient-based methods focus on extracting meta-knowledge required to improve the optimization performance. Model-agnostic meta-learning (MAML)~\cite{DBLP:conf/icml/FinnAL17} regards the initialization of deep network as meta-knowledge and aims to learn a good initialization for all tasks, so that the learner of a new task just needs a few gradient steps from this initialization. R2D2~\cite{DBLP:conf/iclr/BertinettoHTV19} and MetaOpt~\cite{DBLP:conf/cvpr/LeeMRS19} adopt ridge regression or support vector machine~\cite{cortes1995support} as the task-specific learner for each learning task, respectively. With these linear classification models, they can learn a feature embedding model which can generalize well on the new task. Compared with the metric-based method, gradient-based method can be applied to many learning problems, \eg regression and reinforcement learning. However it usually suffers from second-order derivatives, Raghu \etal~\cite{DBLP:conf/iclr/RaghuRBV20} proposed ANIL to relieve this problem.

\subsection{Domain Adaptation}
There is a substantial body of work on domain adaptation~\cite{DBLP:journals/ijon/WangD18}, which aims to learn from one or multiple source domains a well-performing model on a target domain. Early methods of domain adaptation generally rely on instance re-weighting~\cite{DBLP:conf/nips/DudikSP05} or model parameter adaptation~\cite{yang2007cross}. Since the emergence of domain adversarial neural networks
(DANN)~\cite{ganin2016domain}, recent frameworks~\cite{DBLP:conf/iccv/LeeKKJ19,DBLP:conf/iccv/XuLYL19} are mainly based on applying adversarial training to alleviate the domain shift existing in source and target domains. There are also some methods using the discrepancy-based framework to align the marginal distribution between domains~\cite{DBLP:conf/icml/LongZ0J17,DBLP:conf/cvpr/Kang0YH19}.

However, most frameworks on domain adaptation are followed two strict priori, \ie, the label spaces of source and target domains are same and numerous unlabeled images in the target domain are available. According to the argumentations in~\cite{DBLP:conf/iclr/TsengLH020,DBLP:conf/eccv/GuoCKCSSRF20}, these assumptions may not be realistic and restrict the domain adaptation framework to handle novel concepts. Our works consider the scenario of how to improve the generalization of the learning model on the new concept from unseen domains.

\begin{figure*}
  \centering
  \includegraphics[width=14.5cm]{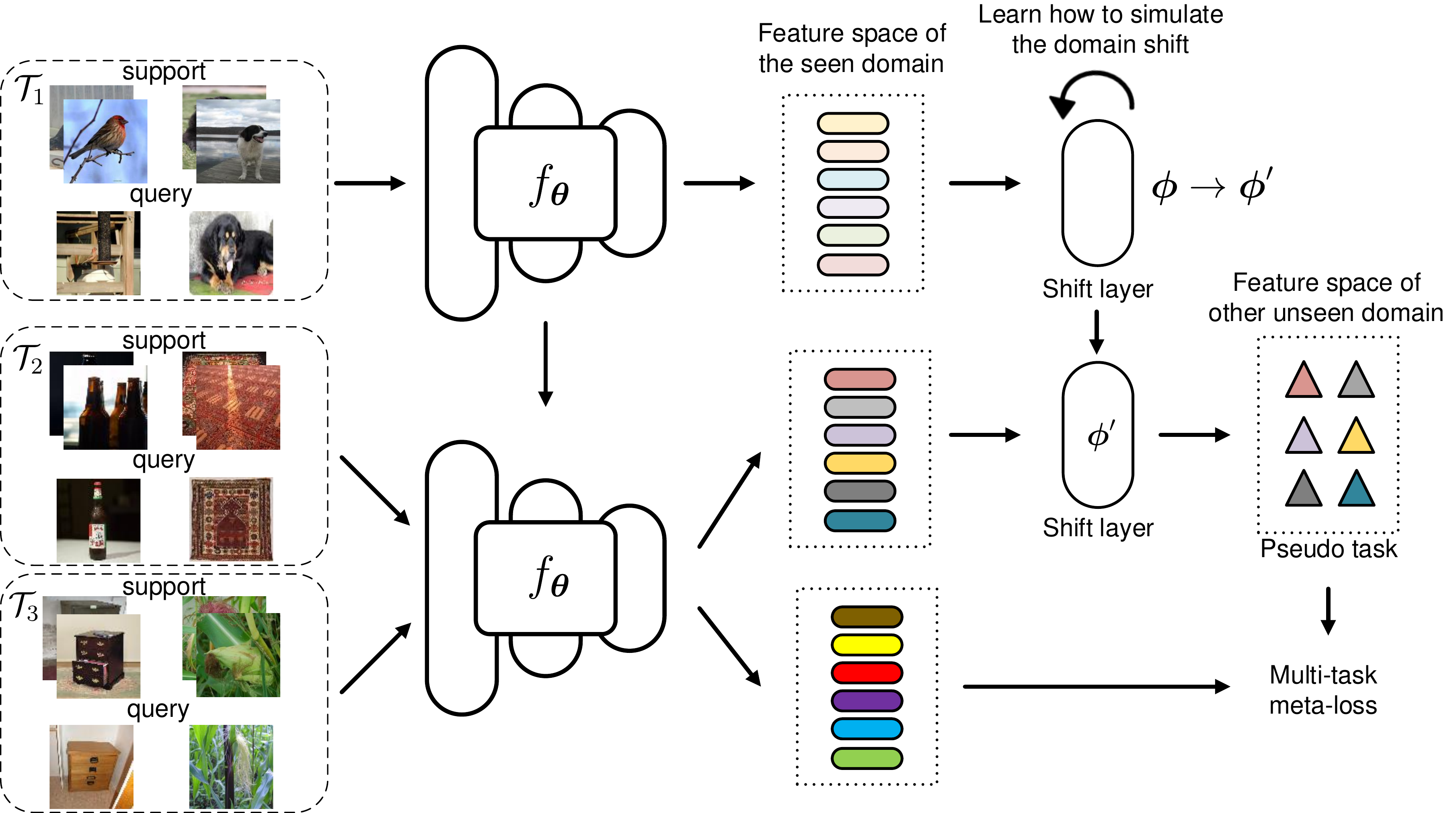}
  \caption{Method overview. In each updating, we sample three tasks from the source domain, \eg, $\mathcal{T}_1$, $\mathcal{T}_2$ and $\mathcal{T}_3$ in this figure. $\mathcal{T}_1$ is used to help the proposed shift layer with the initialization $\boldsymbol \phi$ learn how to simulate the domain shift in unseen domains. Then,  based on $\mathcal{T}_2$, the adapted shift layer $\boldsymbol  \phi'$ can generate a pseudo task $\hat{\mathcal{T}}_2$ that is supposed to be from the other domains. Finally,  $\hat{\mathcal{T}}_2$ and $\mathcal{T}_3$ are used to optimize the meta-parameter together.}   \label{fig:1}
  \vspace{-2mm}
\end{figure*}
\subsection{Domain Generalization}
In contrast to domain adaptation, domain generalization methods~\cite{balaji2018metareg,DBLP:conf/iclr/ShankarPCCJS18,DBLP:journals/corr/abs-2103-01134} devote to learn features that perform well when transferred to unseen domains. These models needn't data from the unseen domains during the training stage. Most recently, meta-learning based approaches~\cite{balaji2018metareg,li2019episodic} are proposed in domain generalization and achieve impressive results. The ideas of these methods are using episodic training to simulate the domain shift during the training and evaluation stages. By this way, better generalization on the unseen domains is achieved. Yet, existing domain generalization approaches still aim at tackling the problem under the assumption that the instances in the training stage share the same label space with the data in the unseen domain.
Besides label space, these algorithms also require to access the training instances drawn from different source domains, not a single one. Our method does not have this limitation.

Compared with domain adaptation and generalization, this work studies a more challenging setting, which just requires one single source domain and needs to generalize well on the new concepts from new domains.

\subsection{Cross-domain Few-shot Classification}
Cross-domain few-shot classification is a scenario which our work can be applied to. In cross-domain few-shot classification, training and novel classes are drawn from different domains, and the class label sets are disjoint. This scenario is very difficult, therefore, limited works aim at cross-domain few-shot classification. Typically, Tsing \etal~\cite{DBLP:conf/iclr/TsengLH020} used feature-wise transform to improve the generalization ability of the learned representations, and Guo \etal~\cite{DBLP:conf/eccv/GuoCKCSSRF20} implemented a broader study of cross-domain few-shot classification and proposed a challenging benchmark. Phoo \etal~\cite{DBLP:journals/corr/abs-2010-07734} introduced a
 self-training approach that allows few-shot learners to adapt feature representations to the
target domain by some unlabeled data from the target domain. LRP~\cite{sun2020explain} uses explanation-guided training to improve the performance.

However, all these approaches are tailored for classification. Specifically, some of them are sensitive to the degree of the domain shift or rely on the transductive setting. Our method is model-agnostic and inductive which can be applied to many learning problems.

\section{Preliminary}
We firstly present the meta-learning problem in the typical few-shot learning, and then generalize to the setting that our method aims to solve.
\subsection{Meta-learning in Typical Few-shot Learning}
In the typical few-shot learning, meta-learning model accesses to a set of training tasks $\mathcal{S}^\text{meta} = \{\mathcal{T}_i\}^T_i$ drawn from a task distribution $P(\mathcal{T})$. Each task $\mathcal{T}_i$ contains a dataset $\mathcal{D}_i$, which is usually divided into two disjoint sets: $\mathcal{D}^\text{tr}_i$ and $\mathcal{D}^\text{ts}_i$. These datasets each is associated with $K$ data-label pairs, \ie, let $\mathbf{x} \in \mathcal{X}$ and $\mathbf{y} \in \mathcal{Y}$ denote data and their label, respectively, $\mathcal{D}^\text{tr}_i = \{
(\mathbf{x}^k_i, \mathbf{y}^k_i)\}^K_{i=1}$, and similarity for $\mathcal{D}^\text{ts}_i$. In general, $\mathcal{D}^\text{tr}_i$ and $\mathcal{D}^\text{ts}_i$ in the same task share the same label space. Different tasks have different label spaces.

We suppose that all the tasks share a same learning algorithm $\mathcal{A}lg$ and a loss function $\mathcal{L}$. Each task $\mathcal{T}_i$  has itself learning model (or base-learner) $\mathcal{A}lg_i$ parameterized by $\mathbf{w}_i \in \mathbb{R}^d $. Meta-learning model is interested in learning a meta-learner, \eg, a neural network, from the meta-training set $\mathcal{S}^\text{meta}$, which can help the base-learner $\mathcal{A}lg_j$ of a new task $\mathcal{T}_{j}$ efficiently learn with a few labeled data. This motivation can be formulated as below:
\begin{align}
   &  \min_{\boldsymbol \theta} \sum_{i=1}^{T} \mathcal{L} ( \boldsymbol \theta, \mathbf{w}_i; \mathcal{D}^\text{ts}_i) \label{eq:bilevel_meta-learner} \\
     & \text{s.t.}~\mathbf{w}_i = \min_{\mathbf{w}} \mathcal{L} (\mathbf{w}; \boldsymbol \theta, \mathcal{D}^\text{tr}_i ) \label{eq:bilelvel_base-learner},
\end{align}
where $\boldsymbol \theta$ denotes the meta-parameter. Particularly, Eq.~\ref{eq:bilevel_meta-learner} and Eq.~\ref{eq:bilelvel_base-learner} can be solved as a bi-level optimization problem~\cite{colson2007overview}.

\textbf{Note:} Metric-based meta-learning method doesn't have the step of Eq.~\ref{eq:bilelvel_base-learner}, because the base-learners of these methods are a nonparametric distance function.

In the meta-test stage, when faced a new task $\mathcal{T}_{j} \in P(\mathcal{T})$, the base-learner $\mathcal{A}lg_j$ can achieve good performance by using the adaptation procedure of $\mathbf{w}_j$ with the learned meta-parameter.

\subsection{Review of Cross-domain Setting}
Different from the typical few-shot learning, we address the few-shot problem under the domain generalization setting. In other words, the meta-training set $\mathcal{S}^\text{meta} = \{\mathcal{T}_i\}^T_{i=i}$ is sampled from a seen (source) domain, and the meta-learning model trained on $\mathcal{S}^\text{meta}$ is supposed to help new tasks from the other unseen domains learn fast. More specifically, there exists a distribution discrepancy between dataset $\mathcal{D}_i$ in the training task $\mathcal{T}_i \in \mathcal{S}^\text{meta}$ and $\mathcal{D}_j$ in the new task $\mathcal{T}_j$. Moreover, we can't access the images in the unseen domain at the training stage.

\section{Methodology}
In this paper, our main idea is to learn how to simulate the domain shift existing in different domains. In this way, we can generate the pseudo tasks from the other domains. With these tasks, the generalization of the meta-learning system on the real unseen domains is supposed to be indeed improved.

\subsection{Feature-wise Shift Layer}
Firstly, we introduce a \textbf{F}eature-w\textbf{i}se \textbf{S}hift \textbf{L}ayer (FiSL) which is used to simulate the domain shift in our method. The architecture of FiSL is based on feature-wise transformation~\cite{DBLP:conf/aaai/PerezSVDC18} which is proven to be capable to represent domain-specific information in many works~\cite{DBLP:conf/iclr/DumoulinSK17,DBLP:conf/aaai/PerezSVDC18}. In few-shot learning, feature-wise transformation is already adopted to dynamically represent domain-specific~\cite{DBLP:conf/iclr/TsengLH020} and task-specific information~\cite{DBLP:conf/nips/OreshkinLL18}.

Suppose that the meta-learner contains a feature encoder $f_{\boldsymbol \theta}$ with the parameter $\boldsymbol \theta \in \Theta$, we are given a feature activation map $\mathbf{z}_0 \in \mathcal{Z}$ of an image $\mathbf{x}_0$ from the last layer of the feature encoder with the dimension of $C \times H \times W$. The output $\mathbf{z}$ of our shift layer is
\begin{equation}\label{eq:shift block}
     \mathbf{z}  = \boldsymbol \gamma \odot \mathbf{z}_0 + \boldsymbol \beta, ~~\text{where}~~\mathbf{z}_0 = f_{\boldsymbol \theta} (\mathbf{x}_0) \in \mathbb{R}^{C \times H \times W},
\end{equation}
$\boldsymbol\gamma$ and $\boldsymbol \beta$ are learnable scaling and shift vectors applied to affine transformation. For easy notation, Eq.~\ref{eq:shift block} can be denoted by $\mathbf{z} = \text{FiSL}(\mathbf{x}_0)$.

After training, the shift layer with $\boldsymbol \phi =\{\boldsymbol \gamma, \boldsymbol \beta\}$ is supposed to be enable to transfer the images from the source domain to other unseen domains.

\subsection{Adversarial Learning-to-learn Mechanism}
However, how to train a meta-learning model with FiSL is an intractable problem because of
\emph{
1. How to make FiSL learn the way of generating pseudo tasks.
2. How to make meta-learning model learn useful domain-invariant meta-knowledge from these tasks.}

\subsubsection{How to Generate Pseudo Tasks}
In the first problem, we are interested in training FiSL in a single domain $P_0$ to simulate the domain shift and generate pseudo tasks from unforeseen domains $P$ for improving the generalization and robustness of meta-learning.

Inspired by the recent developments in robust optimization and adversarial data augmentation~\cite{sinha2017certifiable,DBLP:conf/nips/VolpiNSDMS18}, we consider the first problem following the worst-case problem around the (training) source distribution $P_0$, as
\begin{equation}\label{eq:worst-case}
\begin{split}
  \min_{\boldsymbol \theta} & \sup_{P: D(P,P_0) \leq \rho} \mathbb{E}_P[ \mathcal{L}( \boldsymbol \theta, \mathbf{w}^*; \mathcal{D}^\text{ts}_i)], \\
  & \text{where}~\mathbf{w}^*_i = \mathop{\arg\min}_{\mathbf{w}} \mathcal{L} (\mathbf{w}; \boldsymbol \theta, \mathcal{D}^\text{tr}_i ).
\end{split}
\end{equation}
Here, $P_0$ represents the distribution that images in the seen (source) domain follow. $P$ is the distribution that FiSL simulates. $D(P, P_0)$ is a distance metric on the space of probability distributions. $\mathcal{D}^\text{tr}_i$ and $\mathcal{D}^\text{ts}_i$ are the support (training) and query (test) sets of task $\mathcal{T}_i$ from the source domain $P_0$.


The solution of Eq.~\ref{eq:worst-case} guarantees good performance (robustness) of the learned $\boldsymbol \theta$ against any data distribution $P$ that is $\rho$ distance away from the source domain $P_0$. In other words, meta-parameter $\boldsymbol \theta$ can achieve good generalization on unseen tasks by solving Eq.~\ref{eq:worst-case}.

We firstly focus on how to simulate the unforeseen distributions $P$ by FiSL.
To preserve the semantics of the input samples, similar to~\cite{DBLP:conf/nips/VolpiNSDMS18,DBLP:conf/nips/000300M20}, we use Wasserstein distance defined in the latent feature $\mathcal{Z}$ as our metric $D$ to constrain the distributions FiSL simulates. Let $c_{\boldsymbol \theta}: \mathcal{Z} \times \mathcal{Z} \rightarrow \mathbb{R}_{+} \cup \{\infty\}$ denote the transportation cost of moving mass from $(\mathbf{x}_0, \mathbf{y}_0)$ to $(\mathbf{x}, \mathbf{y})$, as
\begin{equation}\label{eq:wdistance}
  c_{\boldsymbol \theta}((\mathbf{x}_0, \mathbf{y}_0), ( \mathbf{x}, \mathbf{y})) \coloneqq \frac{1}{2} \|\mathbf{z}_0 - \mathbf{z} \|^2_2 + \infty \cdot \mathbf{1}\{\mathbf{y}_0 \neq \mathbf{y}\},
\end{equation}
where $\mathbf{z}_0 = f_{\boldsymbol \theta} (\mathbf{x}_0)$ and $\mathbf{z} = \text{FiSL}(\mathbf{x}_0)$. $\mathbf{x}$ is the pseudo data of $\mathbf{z}$.
For probability measures $P$ and $P_0$ supported on $\mathcal{Z}$, we consider that $\Pi(P, P_0)$ denotes their couplings. Then, the notion of our metric is defined by
\begin{equation}\label{eq:our_wdistance}
  D_{\boldsymbol \theta} (P, P_0) \coloneqq \inf_{M \in \Pi(P, P_0)} \mathbb{E}_M[c_{\boldsymbol \theta}((\mathbf{x}_0, \mathbf{y}_0), (\mathbf{x}, \mathbf{y}))].
\end{equation}
Armed with this notion of distance on the semantic space, we now consider a variant of the worst-case
problem Eq.~\ref{eq:worst-case} where we replace the distance with $D_{\boldsymbol \theta}$ in Eq.~\ref{eq:our_wdistance}, our adaptive notion of distance defined on
the semantic space is
\begin{equation}\label{eq:our_definit}
   \min_{\boldsymbol \theta} \sup_P \{ \mathbb{E}_P[ \mathcal{L}( \boldsymbol \theta, \mathbf{w}^*_i; \mathcal{D}^\text{ts}_i ):  D_{\boldsymbol \theta}(P, P_0)] \leq \rho \}.
\end{equation}

However, for deep neural networks, this formulation is intractable with arbitrary $\rho$. Instead, following the reformulation of~\cite{sinha2017certifiable,DBLP:conf/nips/VolpiNSDMS18}, we
consider its \emph{Lagrangian relaxation}  $\mathcal{F}$ for a fixed penalty parameter $\gamma$
 \begin{equation}\label{eq:sup_max}
 \min_{\boldsymbol \theta} \sup_P \{ \mathbb{E}_P[ \mathcal{L}( \boldsymbol \theta, \mathbf{w}^*_i; \mathcal{D}^\text{ts}_i )- \gamma D_{\boldsymbol \theta}(P, P_0)] \},
\end{equation}
where $\mathbf{w}^*_i = \mathop{\arg\min}_\mathbf{w} = \mathcal{L}(\mathbf{w}; \boldsymbol \theta, \mathcal{D}^\text{tr}_i)$.
Taking the dual reformulation of the penalty relaxation Eq.~\ref{eq:sup_max}, we can obtain an efficient solution
procedure: simulating the unseen distribution $P$ by FiSL, learning the robust $\boldsymbol \theta$ with it.

According to Theorem~\ref{the1} that is a minor adaptation of Lemma 1 in~\cite{DBLP:conf/nips/VolpiNSDMS18}, we propose an iterative training procedure to solve the penalty problem (Eq.~\ref{eq:sup_max}).

\begin{theorem}\label{the1}
Let $ (\Theta \times \mathbb{R}^d) \times (\mathcal{X} \times \mathcal{Y}) \rightarrow R$ and Let $\phi_r$ denote the robust surrogate loss. Then, for any distribution $P_0$ and any $\gamma \geq 0$, we have that,
\begin{equation}
\begin{split}
  & \sup_P  \{ \mathbb{E}_P[ \mathcal{L}( \boldsymbol \theta, \mathbf{w}; \mathcal{D}^\text{ts}_i))- \gamma D_{\boldsymbol \theta}(P, P_0)] \} \\
  & = \mathbb{E}_{(\mathbf{x}, \mathbf{y} )\in \mathcal{D}^\text{ts}_i} [\phi_\gamma(\boldsymbol\theta, \mathbf{w}; \mathbf{x}, \mathbf{y})],~~~~~ \text{where}
\end{split}
\end{equation}
\vspace{-3mm}
\begin{equation}\label{eq:su_loss}
\begin{split}
    & \phi_\gamma(\boldsymbol\theta, \mathbf{w}; \mathbf{x}_0, \mathbf{y}_0) \\
    & =\sup_{\mathbf{x} \in \mathcal{X}} \{ \mathcal{L}( \boldsymbol \theta, \mathbf{w}; \mathbf{x}, \mathbf{y}_0 )- \gamma c_{\boldsymbol \theta}( ( \mathbf{x}_0, \mathbf{y}_0), (\mathbf{x}, \mathbf{y}_0 ) ) \}.
\end{split}
\end{equation}
\end{theorem}
Our training procedure contains two phases: a maximization phase where FiSL learns how to simulate the domain shift by computing the maximization problem (Eq.~\ref{eq:su_loss}) and a minimization phase, where meta-parameter $\boldsymbol \theta$ can perform stochastic gradient descent procedures on the robust surrogate $\phi_\gamma$.
Note that $\mathbf{x}$ in Eq.~\ref{eq:su_loss} is generated by FiSL in our method.

\textbf{Maximization phase.} In the maximization phase, a new task $\mathcal{T}_j$ drawn from the source domain $P_0$ is given to help FiSL learn how to simulate the domain shift. This phase can be formulated as
\begin{equation}\label{eq:max}
\begin{split}
   \boldsymbol \phi' =  \boldsymbol \phi +  \eta \nabla \{ & \mathbb{E}_{(\mathbf{x}_0, \mathbf{y}_0) \in \mathcal{D}^\text{ts}_j}\mathcal{L}(\boldsymbol \theta, \mathbf{w}^*_j; \mathbf{x}, \mathbf{y}_0) \\
   & - \gamma c_{\boldsymbol \theta}(( \mathbf{x}_0, \mathbf{y_0}),( \mathbf{x}, \mathbf{y}_0) )  \},
\end{split}
\end{equation}
where $\mathbf{w}^*_j = \mathop{\arg\min}_\mathbf{w} = \mathcal{L}(\mathbf{w}; \boldsymbol \theta, \mathcal{D}^\text{tr}_j)$ and $\mathbf{x}$ is the pseudo data generated by FiSL.

\textbf{Minimization phase.} With the learned FiSL, the task $\mathcal{T}_i$ can be transformed to a pseudo task $\hat{\mathcal{T}_i}$ from the other domains, which is used to optimize the meta-parameter $\boldsymbol \theta$, such as
\begin{equation}\label{eq:min}
  \boldsymbol \theta = \boldsymbol \theta - \alpha \nabla_{\boldsymbol \theta} \mathcal{L} (\boldsymbol \theta, \mathbf{w}_i^*; \hat{\mathcal{D}}^\text{ts}_i),
\end{equation}
where $\mathbf{w}_i^* = \min_{\mathbf{w}} \mathcal{L} (\mathbf{w}; \boldsymbol \theta, \hat{\mathcal{D}}^\text{tr}_{i} )$, $\hat{\mathcal{D}}^\text{tr}_i = \text{FiSL}_{\boldsymbol \phi'}(\mathcal{D}^\text{tr}_i)$ and $\hat{\mathcal{D}}^\text{ts}_i = \text{FiSL}_{\boldsymbol \phi'}(\mathcal{D}^\text{ts}_i)$.


\subsubsection{How to Learn Domain-invariant Knowledge}
For learning domain-invariant meta-knowledge, inspired by multi-task learning~\cite{zhang2017survey}, besides optimizing $\boldsymbol \theta$ by the pseudo task $\hat{\mathcal{T}}_i$ in Eq.~\ref{eq:min}, we sample an additional task $\mathcal{T}_k$ from the source domain to jointly optimize $\boldsymbol \theta$, such as
\begin{equation}\label{eq:overall-loss}
       \min_{\boldsymbol \theta} \mathcal{L} (\boldsymbol \theta, \mathbf{w}^*_k;  \mathcal{D}^\text{ts}_k)
       +  \mathcal{L} ( \boldsymbol \theta, \mathbf{w}^*_{i} ;  \hat{\mathcal{D}}^\text{ts}_{i}) ,
\end{equation}
where $\mathbf{w}^*_{k}  = \min_{\mathbf{w}} \mathcal{L} (\mathbf{w}; \boldsymbol \theta, \mathcal{D}^\text{tr}_{k})$ and $\mathbf{w}_{i}^*  = \min_{\mathbf{w}} \mathcal{L} (\mathbf{w}; \boldsymbol \theta, \hat{\mathcal{D}}^\text{tr}_{i})$. $\hat{\mathcal{D}}^\text{tr}_{i}$ and $\hat{\mathcal{D}}^\text{ts}_{i}$ is transformed by FiSL learned by Eq.~\ref{eq:max}.

Moreover, for learning cross-domain meta-knowledge, FiSL is supposed to dynamically generate various unseen domains based on different tasks. Hence, similar to MAML, we learn a good initialization for FiSL. From the learned initialization, an unseen domain can be simulated by a few gradient descents by Eq.~\ref{eq:max}. In particular, the good initialization is appropriate for simulating many unseen domains, in the sense that the learned initialization can be regarded as including domain-invariant knowledge. In the meta-test stage, we can transform the data from unseen domains by this initialization to achieve better generalization. The full algorithm is summarized in Algorithm~\ref{alg1}. The overview of our method can be found in Figure.~\ref{fig:1}.
\begin{algorithm}\label{alg1}
\caption{Learning-to-learn Adversarial Shift}
\SetAlgoLined
\KwIn{ Sample three sets of training tasks $\mathcal{S}_1 = \{\mathcal{T}_{3i-2}\}_{i = 1}^N, \mathcal{S}_2 = \{\mathcal{T}_{3i-1}\}_{i = 1}^{N}$ and $\mathcal{S}_3 = \{\mathcal{T}_{3i}\}_{i = 1}^{N}$  }
\KwOut {Learned weights $\boldsymbol \theta, \boldsymbol \phi$}
Initialize $\boldsymbol \theta, \boldsymbol \phi$ \\
 \While{not converged}{
   \For { i = 1, \ldots, N}
   {
      Train a base-learner $\mathbf{w}^*_{3i-2}$ for $\mathcal{T}_{3i-2}$ with $\mathcal{D}^\text{tr}_{3i-2}$ \\
      Use $\mathcal{T}_{3i-2}$ to obtain $\boldsymbol \phi'$ via Eq.~\ref{eq:max} \label{step:5}  \\
      Generate a pseudo tasks $\hat{\mathcal{T}}_{3i-1}$ based on $\mathcal{T}_{3i-1}$, $ \hat{\mathcal{D}}^\text{tr}_{3i-1} = \text{FiSL}_{\boldsymbol \phi'} (\mathcal{D}^\text{tr}_{3i-1})$ and $ \hat{\mathcal{D}}^\text{ts}_{3i-1} = \text{FiSL}_{\boldsymbol \phi'} (\mathcal{D}^\text{ts}_{3i-1})$ \\
      Train base-learners for pseudo task $\hat{\mathcal{T}}_{3i-1}$ and task $\mathcal{T}_{3i}$, respectively.\\
      Update $\boldsymbol \theta, \boldsymbol \phi$ via multi-task loss in Eq.~\ref{eq:overall-loss}
      \vspace{-3mm}
      \begin{flalign}
       & \boldsymbol \theta \leftarrow   \boldsymbol \theta -\alpha \nabla_{\boldsymbol \theta} [  \mathcal{L} (\boldsymbol \theta, \mathbf{w}^*_{3i} ; \mathcal{D}^\text{ts}_{3i}) & \nonumber \\
           & ~~~~~~+ \mathcal{L} (\boldsymbol \theta, \mathbf{w}^*_{3i-1} ; \hat{\mathcal{D}}^\text{ts}_{3i-1})] & \nonumber \\
       & \boldsymbol \phi \leftarrow   \boldsymbol \phi -\alpha \nabla_{\boldsymbol \phi} [  \mathcal{L} (\boldsymbol \theta, \mathbf{w}^*_{3i} ; \mathcal{D}^\text{ts}_{3i}) & \nonumber \\
           & ~~~~~~+ \mathcal{L} (\boldsymbol \theta, \mathbf{w}^*_{3i-1} ; \hat{\mathcal{D}}^\text{ts}_{3i-1})] & \nonumber
      \end{flalign}
      \vspace{-7mm}
   }
 }
\end{algorithm}

\textbf{Note:} When Algorithm~\ref{alg1} is applied to metric-based meta-learning method, it's not necessary to train a base learner of each task, \ie, step 4 and 7.

In the meta-test stage, the learned $\boldsymbol \theta^*$ and $\boldsymbol \phi^*$ can help meta-learning method achieve good generalization on the new task $\mathcal{T}_l$ from the unseen domains, as
\begin{equation}\label{eq:meta-test}
  \mathbf{w}_l = \min_{\mathbf{w}} \mathcal{L} (\mathbf{w}; \boldsymbol \theta^*, \hat{\mathcal{D}}^\text{tr}_l), \hat{\mathcal{D}}^\text{tr}_l = \text{FiSL}_{\boldsymbol \phi^*}(\mathcal{D}^\text{tr}_l).
  \vspace{-2mm}
\end{equation}

\begin{figure}
  \centering
  \includegraphics[width=8.5cm]{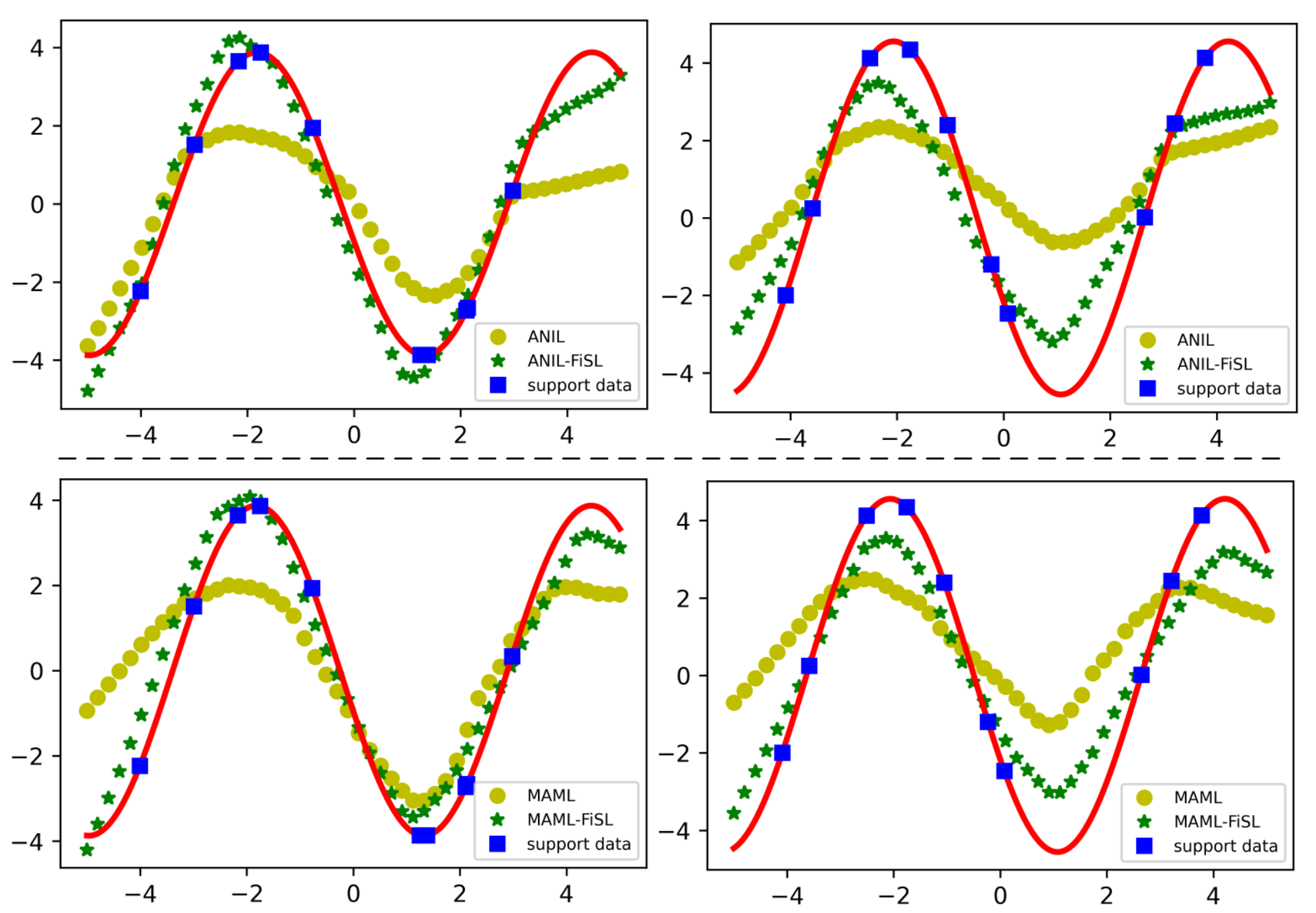}
  \caption{Results of cross-domain few-shot regression. Top: results of ANIL and ANIL-FiSL (our). Below: results of MAML and MAML-FiSL (our). Redline is the ground truth. Our method can help different meta-learning models achieve better performance on the unseen domain. }\label{fig:2}
  \vspace{-2mm}
\end{figure}
\section{Experiments}
In this section, we evaluate our method in different learning problems, including regression and classification to verify the adaptation of the proposed method. In each setting, our method is used to improve the generalization of different metric-based and gradient-based meta-learning models on unseen domains to demonstrate that our method is  model-agnostic and can indeed help learn robust meta-knowledge.

\subsection{Cross-domain Regression}
\textbf{Experimental Setup.} We start with a regression problem of fitting sine curves following~\cite{DBLP:conf/icml/FinnAL17}. Each task involves regressing from the input to the output of a sine wave.
The amplitude and phase of the training tasks are uniformly sampled from $[ 0.1, 3.0]$ and $[0, \frac{3}{4}\pi]$, respectively. However, the amplitude and phase of test tasks are uniformly drawn from $[ 3.0, 5.0]$ and $[\frac{3}{4}\pi, \pi]$, respectively. For training, five labeled datapoints are given for each task as $\mathcal{D}^\text{tr}$, and twenty labeled datapoints are sampled as $\mathcal{D}^\text{ts}$. Both of them are uniformly sampled from $[-5, 5]$. We use a neural network with two hidden layers as the feature encoder and
40 nodes each and a mean-squared error (MSE) as loss $\mathcal{L}$.

Our method is applied to two gradient-based methods: MAML~\cite{DBLP:conf/icml/FinnAL17} and ANIL~\cite{DBLP:conf/iclr/RaghuRBV20}. Because, our method is trained by multi-task mechanism, for fairness, the batchsize of tasks to train MAML and ANIL is set to 2. All models are trained $20,000$ iterations by Adam with a learning rate of 0.001. Both MAML and ANIL use one inner gradient step with a learning rate 0.01 and 0.1, respectively. During the test, we present the model with $2,000$ newly sampled tasks from the disjoint domain and measure mean squared error over 100 test points on each task. $\gamma$ and $\eta$ in our method are $0.5$ and $0.01$, respectively.

\textbf{Results.} Table~\ref{tab:reg} shows the results of different models on the cross-domain few-shot regression. We can see that our method effectively improves the generalization of different gradient-based models on the unseen domain. The results also verify that our method can work well in regression problem. Figure.~\ref{fig:2} represents some results of these models on some new tasks from the unseen domain.

\begin{table}
\begin{center}
\scalebox{0.8}{
\begin{tabular}{c|c|cc}
\toprule
 Methods  & FiSL&  5-shot & 10-shot   \\
 \midrule
  \multirow{2}{4em}{ANIL~\cite{DBLP:conf/iclr/RaghuRBV20}} & - &	4.256 $\pm$ 0.127 & 3.080 $\pm$ 0.075  \\
&  $\surd$ & \textbf{1.889 $\pm$ 0.075} &	\textbf{0.961 $\pm$ 0.034} \\
\midrule
\multirow{2}{5em}{MAML~\cite{DBLP:conf/icml/FinnAL17}} &- & 3.558 $\pm$ 0.087 & 2.168 $\pm$ 0.060 \\
 & $\surd$ & \textbf{1.712 $\pm$ 0.075} & \textbf{0.935 $\pm$ 0.035} \\
  \bottomrule

\end{tabular}}
\end{center}
 \caption{Mean squared error (MSE) of cross-domain few-shot regression, lower is better. FiSL indicates that we apply the shift layer with the proposed adversarial mechanism to train the model.}
\label{tab:reg}
\vspace{-2mm}
\end{table}

\subsection{Cross-domain Classification}

\begin{figure}
  \centering
  \includegraphics[width=7cm]{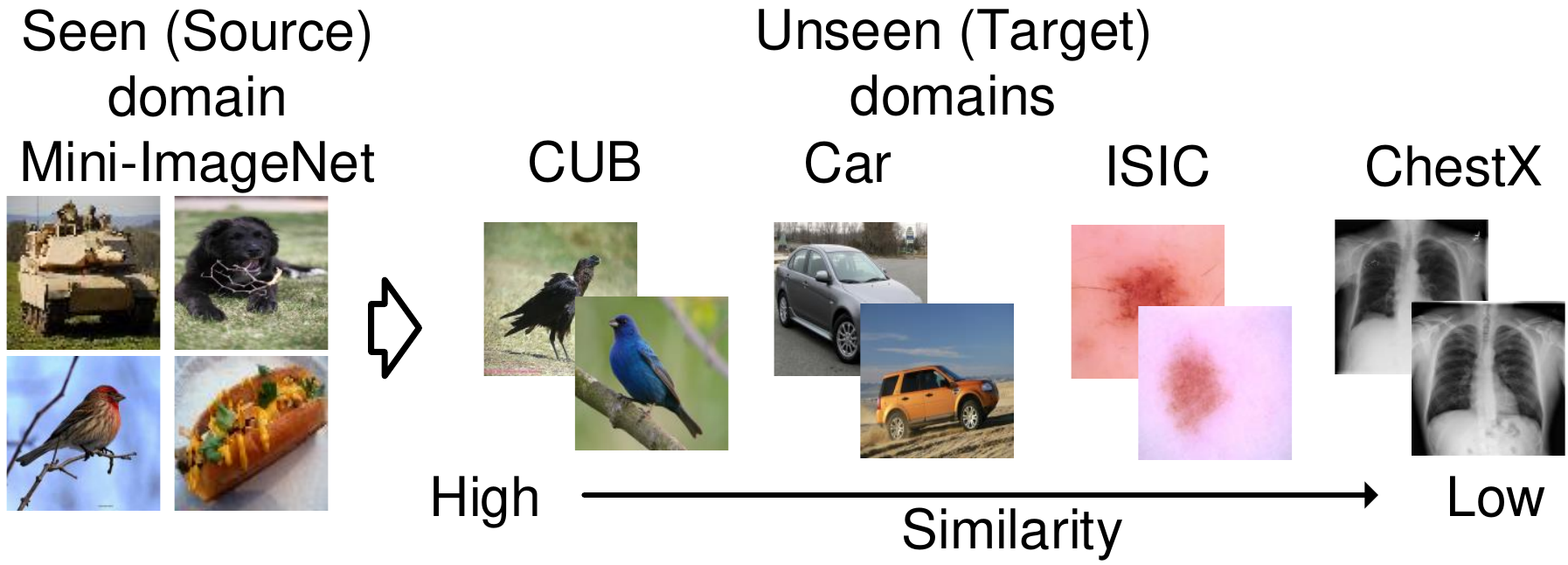}
  \caption{Images from different benchmarks. mini-ImageNet is used as the source domain, and domains of varying dissimilarity from natural images are used for target evaluation.}\label{fig:3}
  \vspace{-3mm}
\end{figure}

\begin{table*}
\begin{center}
\scalebox{0.85}{
\begin{tabular}{c|c|cccc}
\toprule
 5way 1-shot& FiSL & CUB & Car & ISIC & ChestX \\
 \midrule
 \multirow{2}{4em}{ProNet} & - & 40.03 $\pm$ 0.58 &	30.60 $\pm$ 0.48 &		30.63 $\pm$ 0.47 &	22.20 $\pm$ 0.33  \\
  & $\surd$ & \textbf{40.94 $\pm$ 0.58} &	\textbf{31.11 $\pm$ 0.48}	&	30.52 $\pm$ 0.47 &	\textbf{22.43 $\pm$ 0.35}\\
 \midrule
\multirow{2}{4em}{RelationNet} & - & 39.30 $\pm$ 0.56	& 28.34 $\pm$ 0.43 &	29.64 $\pm$ 0.46 &	22.12 $\pm$ 0.33\\
   & $\surd$ & \textbf{40.29 $\pm$ 0.57}	& \textbf{29.00 $\pm$ 0.46}	&	28.75 $\pm$ 0.44  &	21.92 $\pm$ 0.32\\
 \midrule
\multirow{2}{4em}{ANIL}  & - & 32.11 $\pm$ 0.55 &	26.58 $\pm$ 0.41 &		24.45 $\pm$ 0.34 &	21.19 $\pm$ 0.25\\
  & $\surd$ & \textbf{39.49 $\pm$ 0.59}	& \textbf{30.21 $\pm$ 0.50}	&	\textbf{31.09 $\pm$ 0.47}	& \textbf{21.88 $\pm$ 0.32} \\
 \midrule
\multirow{2}{4em}{MetaOpt} & - &42.33 $\pm$ 0.60 &	32.04 $\pm$ 0.47 &		29.61 $\pm$ 0.45 &	22.10 $\pm$ 0.31\\
   & $\surd$ & \textbf{45.53 $\pm$ 0.61} &	\textbf{34.67 $\pm$ 0.51} &		\textbf{33.82 $\pm$ 0.51} &	\textbf{22.91 $\pm$ 0.33} \\
 \midrule
\multirow{2}{4em}{R2D2} & - & 42.81 $\pm$ 0.62 &	33.15 $\pm$ 0.49 &		30.04 $\pm$ 0.46 &	 22.35 $\pm$ 0.32 \\
  & $\surd$ & \textbf{44.14 $\pm$ 0.60}	& \textbf{34.18 $\pm$ 0.51}		& \textbf{32.47 $\pm$ 0.49}	& \textbf{23.02 $\pm$ 0.34}\\
  \midrule
\midrule
 5-way 5-shtot & FiSL & CUB & Car & ISIC & ChestX \\
 \midrule
 \multirow{2}{4em}{ProNet} & - & 57.26$\pm$ 0.57 &	42.83 $\pm$ 0.53 &	39.89 $\pm$ 0.43 & 25.03 $\pm$ 0.34  \\
  & $\surd$ & \textbf{57.84 $\pm$ 0.56} & \textbf{43.11  $\pm$ 0.54} & \textbf{41.47 $\pm$ 0.43} & \textbf{25.40 $\pm$0.35} \\
 \midrule
 \multirow{2}{4em}{RelationNet} & - & 55.71 $\pm$ 0.56 &	37.86 $\pm$ 0.51 &	38.10 $\pm$  0.43 & 24.11 $\pm$  0.32 \\
  & $\surd$ & \textbf{56.11 $\pm$ 0.53} &	\textbf{39.66 $\pm$ 0.55} &		\textbf{38.39 $\pm$ 0.45}	& 23.73 $\pm$ 0.32\\
 \midrule
 \multirow{2}{4em}{ANIL} & - & 37.24 $\pm$ 0.57 &	28.79 $\pm$ 0.40 &	27.90 $\pm$ 0.38 &	20.93 $\pm$ 0.17\\
  & $\surd$ & \textbf{57.56 $\pm$ 0.58} &	\textbf{44.34 $\pm$ 0.57}	&	\textbf{41.82 $\pm$ 0.45} & \textbf{25.00 $\pm$ 0.34}\\
 \midrule
 \multirow{2}{4em}{MetaOpt} & - & 61.66  $\pm$  0.60 &	50.55  $\pm$ 0.56 &		44.10  $\pm$  0.47 &	26.36  $\pm$  0.35\\
  & $\surd$ &  \textbf{63.24  $\pm$ 0.57} &	 \textbf{50.82  $\pm$ 0.58}	&	 \textbf{46.39  $\pm$ 0.46} &	 \textbf{26.46  $\pm$ 0.34} \\
 \midrule
 \multirow{2}{4em}{R2D2} & - & 62.31 $\pm$ 0.59 &	49.49 $\pm$ 0.57 &		42.81 $\pm$ 0.44 &	25.92 $\pm$ 0.34\\
  & $\surd$ &  \textbf{64.62 $\pm$ 0.58}	&  \textbf{51.62 $\pm$ 0.60} &		 \textbf{47.62 $\pm$ 0.47} &	 \textbf{26.48 $\pm$ 0.36}\\
\bottomrule
\end{tabular}}
\end{center}
\caption{Few-shot classification results on unseen domains. We train the model on the mini-ImageNet domain and evaluate the trained model on other domains. FiSL is our method.}
\label{tab:results_metamodels}
\vspace{-2mm}
\end{table*}
In cross-domain few-shot classification, we validate the efficacy of the proposed methods with two categories of meta-learning frameworks, \ie, metric-based and gradient-based frameworks. In metric-based method, we choose ProNet~\cite{DBLP:conf/nips/SnellSZ17} and RelationNet~\cite{DBLP:conf/cvpr/SungYZXTH18}. As for gradient-based method, ANIL, R2D2~\cite{DBLP:conf/iclr/BertinettoHTV19}, and MetaOptNet~\cite{DBLP:conf/cvpr/LeeMRS19} are chosen. In order to evaluate the performance on unseen domains, we
train the few-shot classification model on the mini-ImageNet~\cite{DBLP:conf/nips/VinyalsBLKW16} domain and evaluate the trained model on four different domains: CUB~\cite{wah2011caltech}, Cars~\cite{krause20133d}, ISIC~\cite{tschandl2018ham10000}, and ChestX~\cite{wang2017chestx}. CUB and Car are two benchmarks which are well established for cross-domain few-shot classification.
Evaluating on these two benchmarks can provide a fair comparison to the previous methods. However, the images of these two datasets are natural images that still retain a high degree of visual similarity. Moreover, according to~\cite{DBLP:conf/eccv/GuoCKCSSRF20}, some previous state-of-the-art methods, \eg~\cite{DBLP:conf/iclr/TsengLH020} are not robust to the large domain shift. Therefore, following~\cite{DBLP:conf/eccv/GuoCKCSSRF20}, we adopt ISIC and Chexst as the other two benchmarks. Some images from these datasets are shown in Figure.~\ref{fig:3}.

\textbf{DataSets.} We conduct experiments using five datasets: mini-ImageNet,
CUB, Cars, ISIC, and ChestX. We follow the same dataset processing in~\cite{DBLP:conf/eccv/GuoCKCSSRF20,DBLP:conf/iclr/TsengLH020}. Compared with natural images in CUB and Car, ISIC and ChestX cover dermoscopic images of skin lesions, and X-ray images respectively, which are largely different to mini-ImageNet. Similar to~\cite{DBLP:conf/iclr/TsengLH020}, we select the training iterations with the best accuracy on the validation set of mini-ImageNet for evaluation.

\begin{table*}
\begin{center}
\scalebox{0.8}{
\begin{tabular}{c|cc|cccc}
\toprule
Method& Pre-trained& FiSL & CUB & Car & ISIC & ChestX  \\
 \midrule
   \multirow{3}{4em}{ProNet} &  &  & 51.82 $\pm$ 0.58 &	42.12 $\pm$ 0.56 & 	39.41 $\pm$ 0.43 &	25.11 $\pm$ 0.35 \\
 & $\surd$ & & 57.26$\pm$ 0.57 &	42.83 $\pm$ 0.53 &	39.89 $\pm$ 0.43 & 25.03 $\pm$ 0.34 \\
 & $\surd$ & $\surd$ & \textbf{57.84 $\pm$ 0.56} & \textbf{43.11  $\pm$ 0.54} & \textbf{41.47 $\pm$ 0.43} & \textbf{25.40 $\pm$0.35} \\
 \midrule
 \multirow{3}{4em}{R2D2}& & & 61.19 $\pm$  0.60 &	43.91 $\pm$  0.59 &	40.57 $\pm$  0.43 &	25.10 $\pm$  0.32 \\
  & $\surd$ & & 62.31 $\pm$ 0.59 &	49.49 $\pm$ 0.57 &		42.81 $\pm$ 0.44 &	25.92 $\pm$ 0.34 \\
 & $\surd$ & $\surd$ & \textbf{64.62 $\pm$ 0.58}	&  \textbf{51.62 $\pm$ 0.60} &		 \textbf{47.62 $\pm$ 0.47} &	 \textbf{26.48 $\pm$ 0.36} \\
\bottomrule
\end{tabular}}
\end{center}
\caption{$5$-way $5$-shot classification accuracy on different unseen domains. This result shows the influence of pre-training on generalization of different meta-learning methods. }
\label{tab:pretrain}
\vspace{-2mm}
\end{table*}

\textbf{Implementation details.}
All experiments are performed with ResNet-10~\cite{DBLP:conf/cvpr/HeZRS16} for a fair comparison. Same as~\cite{DBLP:conf/eccv/GuoCKCSSRF20,DBLP:conf/iclr/TsengLH020}, we firstly pre-train ResNet-10 by minimizing the standard cross-entropy classification loss on the $64$ training categories in the mini-ImageNet dataset. Then all the models are trained $30,000$ iterations by Adam~\cite{DBLP:journals/corr/KingmaB14} with a learning rate of $0.001$. Because our method uses multi-task training, the batchsize is set to $2$ for training ProNet, RelationNet, ANIL, R2D2, and MetaOpt. The inner learning rate and updating steps are $0.1$ and $5$ for ANIL. $\gamma$ and $\eta$ in our method are $0.5$ and $0.1$, respectively.

We present the average results over $1,000$ trials for all the experiments, and report the average accuracy and $95\%$ confidence interval. In each trial, the query set $\mathcal{D}^\text{ts}$ contains $15$ images.

\begin{table}
\begin{center}
\scalebox{0.8}{
\begin{tabular}{c|c|cc}
\toprule
Method& Shot & CUB & Car  \\
 \midrule
 $\text{GNN}^\dag$~\cite{DBLP:conf/iclr/SatorrasE18} & 1 & 45.69 $\pm$ 0.68	&  31.79 $\pm$ 0.51\\
 GNN-FWT~\cite{DBLP:conf/iclr/TsengLH020} & 1 & \textbf{47.47 $\pm$ 0.75}	&  31.61 $\pm$ 0.53 \\
 LRP-CAN~\cite{sun2020explain} & 1 & 46.23 $\pm$ 0.42	&  32.66 $\pm$ 0.46 \\
 \midrule
 \textbf{R2D2-FiSL}& 1 & 44.14 $\pm$ 0.60	&  34.18 $\pm$ 0.51 \\
 \textbf{MetaOpt-FiSL} & 1 & 45.53 $\pm$ 0.60	&  \textbf{34.67 $\pm$ 0.51} \\
 \midrule
 \midrule
  $\text{GNN}^\dag$~\cite{DBLP:conf/iclr/SatorrasE18} & 5 & 62.25 $\pm$ 0.65	&  44.28 $\pm$ 0.63\\
 GNN-FWT~\cite{DBLP:conf/iclr/TsengLH020} & 5 & \textbf{66.98 $\pm$ 0.68}	&  44.90 $\pm$ 0.64 \\
 LRP-CAN~\cite{sun2020explain} & 5 & 66.58 $\pm$ 0.39	&  43.86 $\pm$ 0.38 \\
 \midrule
 \textbf{MetaOpt-FiSL} & 5 & 63.24 $\pm$ 0.57	&  50.82 $\pm$ 0.58 \\
 \textbf{R2D2-FiSL} & 5 & 64.62 $\pm$ 0.58	&  \textbf{51.62 $\pm$ 0.60} \\
\bottomrule
\end{tabular}}
\end{center}
\caption{$5$-way $K$-shot classification accuracy on CUB and Car. $^\dag$ Results reported in~\cite{DBLP:conf/eccv/GuoCKCSSRF20}.}
\label{tab:results_SOTA}
\vspace{-2.5mm}
\end{table}

\textbf{Generalization with FiSL.} Table~\ref{tab:results_metamodels} shows the results of five meta-learning models with FiSL or not on $5$way-$1$ shot and $5$way-$5$shot cross-domain few-shot classification. Both the metric-based and gradient-based models trained with our method perform favorably against the individual baselines. This observation demonstrates that our method is model agnostic and robust to different unseen domains of varying dissimilarity from the source domain. We attribute the improvement of the generalization to the use of FiSL for making the meta-learner really learn the domain-invariant meta-knowledge.

We also obverse that gradient-based methods achieve better generalization than metric-based methods. This might due to the gradient-based methods learn a base learner for the new task with the provided labeled data. The defined metric space learned in the source domain by metric-based methods is not flexible compared with adapting learner on unseen domains.

\begin{table}
\begin{center}
\scalebox{0.8}{
\begin{tabular}{c|c|cc}
\toprule
Method& Shot & ISIC & ChestX  \\
 \midrule
 $\text{ProNet}^\dag$ & 5 & 39.57 $\pm$ 0.57	&  24.05 $\pm$ 1.01\\
 $\text{ProNet-FWT}^\dag$~\cite{DBLP:conf/iclr/TsengLH020} & 5 & 38.87 $\pm$ 0.52	&  23.77 $\pm$ 0.42 \\
 $\text{RN}^\dag$ & 5 & 39.41 $\pm$ 0.58	&  22.96 $\pm$ 0.88 \\
 $\text{RN-FWT}^\dag$~\cite{DBLP:conf/iclr/TsengLH020} & 5 & 35.54 $\pm$ 0.55	&  22.74 $\pm$ 0.40 \\
 $\text{MAML}^\dag$ & 5 & 40.13 $\pm$ 0.58	&  23.48 $\pm$ 0.96\\
 CHEF~\cite{adler2020cross} & 5 & 41.26 $\pm$ 0.34	&  24.72 $\pm$ 0.14\\
  $\text{Fixed}^\dag$~\cite{DBLP:conf/eccv/GuoCKCSSRF20} & 5& 43.56 $\pm$ 0.60	&  25.35 $\pm$ 0.96\\
 \midrule
  \textbf{MetaOpt-FiSL} & 5 & 46.39 $\pm$ 0.46	&  26.46 $\pm$ 0.34 \\
 \textbf{R2D2-FiSL}& 5 & \textbf{47.62 $\pm$ 0.47}	&  \textbf{26.48 $\pm$ 0.36} \\
 \midrule
 \midrule
  $\text{ProNet-FWT}^\dag$~\cite{DBLP:conf/iclr/TsengLH020} & 20 & 43.78 $\pm$ 0.47	&  26.87 $\pm$ 0.43\\
 $\text{RN-FWT}^\dag$~\cite{DBLP:conf/iclr/TsengLH020} & 20 & 43.31 $\pm$ 0.51	&  26.75 $\pm$ 0.41 \\
 CHEF~\cite{adler2020cross} & 20 & 54.30 $\pm$ 0.34	&  29.71 $\pm$ 0.27 \\
 $\text{Fixed}^\dag$~\cite{DBLP:conf/eccv/GuoCKCSSRF20} & 20 & 52.78 $\pm$ 0.39	&  30.83 $\pm$ 1.05 \\
 \midrule
 \textbf{MetaOpt-FiSL} & 20 & 55.34 $\pm$ 0.44	&  30.59 $\pm$ 0.35 \\
 \textbf{R2D2-FiSL} & 20 & \textbf{58.74 $\pm$ 0.46}	&  \textbf{31.51 $\pm$ 0.36} \\
\bottomrule
\end{tabular}}
\end{center}
\caption{$5$-way $K$-shot classification accuracy on ISIC and ChestX. RN indicates RelationNet. Fixed (Fixed feature extractor) in~\cite{DBLP:conf/eccv/GuoCKCSSRF20} is a strong baseline that leverages the pre-trained model as a fixed
feature extractor and a linear model as the classifier. Many meta-learning models can't outperform it when exists a large domain shift. $^\dag$ Results reported in~\cite{DBLP:conf/eccv/GuoCKCSSRF20}.}
\label{tab:results_SOTA_2}
\vspace{-3mm}
\end{table}

\textbf{Comparison to previous state of the arts.} Table~\ref{tab:results_SOTA} and
Table~\ref{tab:results_SOTA_2} show the results. All the models are trained on mini-ImageNet and evaluate on the other four benchmarks. First, we obverse GNN+FWT achieves the best performance on CUB, however, degrades on Car. Meanwhile, our method achieves competitive performance on CUB and outperforms GNN-FWT $3.06\%$ and $6.72\%$ on 1-shot and 5-shot on Car, respectively. We think that CUB is a fine-grained bird dataset, which has the highest similarity to the mini-ImageNet in semantic content and color style among these four datasets, as shown in Figure.~\ref{fig:3}. Some recent few-shot learning methods~\cite{DBLP:conf/eccv/AfrasiyabiLG20} can even handle this situation with a small domain shift. From this point, learning well enough on mini-ImageNet could provide satisfying performance on CUB.

On ISIC and ChestX, our method achieves state-of-the-art performance and largely outperforms the previous methods. We also obverse FWT can't improve the performance of ProNet and RelationNet. Some similar results of our method can be found in Table~\ref{tab:results_metamodels}. Note that the decline of our methods is less than FWT. In this regard, it's challenging to improve the generalization of the metric-based meta-learning on some domains existing a large shift.

Moreover, some recent works~\cite{tian2020rethinking,chen2019closer} point out that meta-learning based few-shot learning algorithms underperform compared to the traditional
pre-training model when there exists a large shift between base and novel class domains. Comparing the performance of MAML, ProNet with Fixed, we can find the same results in Table~\ref{tab:results_SOTA_2}. However, based on our method, some meta-learning methods can largely outperform Fixed.

\subsection{Influence of Pre-training on the Generalization}
According to several recent methods~\cite{DBLP:conf/cvpr/YeHZS20,DBLP:conf/iclr/RusuRSVPOH19}, pre-training can significantly improve the performance of meta-learning frameworks on the typical few-shot learning scenario. In this section, we investigate the influence of pre-training on the generalization on unseen domains.

As shown in Table~\ref{tab:pretrain}, pre-training the feature encoder substantially improves the performance of ProNet and R2D2 on four unseen benchmarks. However, the influence of pre-training on ProNet is not very obvious, when there exists a large domain shift in target domains.

\section{Conclusion}

We propose a model-agnostic method to effectively enhance the generalization of different kinds of meta-learning frameworks under the
domain shift, which can be applied to many learning problems. The core idea of our method lies in using the feature-wise shift layer to
simulate various distributions existing in unseen domains. In order to learn how to simulate the distributions and learn domain-invariant knowledge,
We develop a learning-to-learn approach for jointly optimizing the proposed feature-wise shift
layer and the meta-learning model. From extensive
experiments, we demonstrate that our method is applicable to different meta-learning frameworks and shows consistent improvement over the baselines
and robustness to different unseen domains.
{\small
\bibliographystyle{ieee_fullname}
\bibliography{egbib}
}

\end{document}